\documentclass[conference]{IEEEtran}
\IEEEoverridecommandlockouts

\usepackage{cite}
\usepackage{amsmath,amssymb,amsfonts}
\usepackage{algorithmic}
\usepackage{graphicx}
\usepackage{textcomp}
\usepackage{xcolor}
\usepackage{setspace}
\usepackage{listings}
\usepackage{glossaries}
\usepackage{microtype}
\usepackage{multirow}
\usepackage{multicol}
\usepackage[most]{tcolorbox}
\usepackage{makecell}
\usepackage{todonotes}
\usepackage{hyperref}
\usepackage{orcidlink}

\newcommand{\llm}{LLM}
\newcommand{\approach}{\texttt{SHERPA}}
\newcommand{\direct}{\texttt{Direct}}

\newcommand{\set}[1]{\mathcal{#1}}

\newcommand{\fscore}{$F_1$}
\newcommand{\recall}{\textit{R}}
\newcommand{\precision}{\textit{P}}
\newcommand{\pass}{\textit{Pass@1}}
\newcommand{\accuracy}{\textit{ACC}}

\def\BibTeX{{\rm B\kern-.05em{\sc i\kern-.025em b}\kern-.08em
    T\kern-.1667em\lower.7ex\hbox{E}\kern-.125emX}}
\begin{document}

\title{\texttt{SHERPA}: A Model-Driven Framework for \\Large Language Model Execution

}

\author{
\IEEEauthorblockN{Boqi Chen\orcidlink{0000-0002-1451-3603}\IEEEauthorrefmark{1},
Kua Chen\orcidlink{0009-0002-7491-7084}\IEEEauthorrefmark{1},
Jos\'e Antonio Hern\'andez L\'opez\orcidlink{0000-0003-2439-2136}\IEEEauthorrefmark{2}\\
Gunter Mussbacher\orcidlink{0009-0006-8070-9184}\IEEEauthorrefmark{3}
D\'aniel Varr\'o\orcidlink{0000-0002-8790-252X}\IEEEauthorrefmark{3}\IEEEauthorrefmark{4},
Amir Feizpour\IEEEauthorrefmark{5}
}
\IEEEauthorblockA{\IEEEauthorrefmark{1}\textit{Electrical and Computer Engineering, McGill University, Canada - \{firstname.lastname\}@mail.mcgill.ca}\\
\IEEEauthorrefmark{2}\textit{Department of Computer Science and Systems, University of Murcia, Spain - joseantonio.hernandez6@um.es}  \\
\IEEEauthorrefmark{3}\textit{Electrical and Computer Engineering, McGill University, Canada - \{firstname.lastname\}@mcgill.ca} \\
\IEEEauthorrefmark{4}\textit{Department of Computer and Information Science, Link\"oping University, Sweden - daniel.varro@liu.se} \\
\IEEEauthorrefmark{5}\textit{Aggregate Intellect Inc., Canada - amir@ai.science} 
}
\thanks{
Partially supported by the FRQNT-B2X project (file number: 319955), IT30340 Mitacs Accelerate, and the Wallenberg AI, Autonomous Systems
and Software Program (WASP), Sweden
}
}

\maketitle

\begin{abstract}
Recently, large language models (LLMs) have achieved widespread application across various fields. Despite their impressive capabilities, LLMs suffer from a lack of structured reasoning ability, particularly for complex tasks requiring domain-specific best practices, which are often unavailable in the training data. Although multi-step prompting methods incorporating human best practices, such as chain-of-thought and tree-of-thought, have gained popularity, they lack a general mechanism to control LLM behavior.
In this paper, we propose \approach{}, a model-driven framework to improve the LLM performance on complex tasks by explicitly incorporating domain-specific best practices into hierarchical state machines. By structuring the LLM execution processes using state machines, \approach{} enables more fine-grained control over their behavior via rules or decisions driven by machine learning-based approaches, including LLMs.
We show that \approach{} is applicable to a wide variety of tasks---specifically, code generation, class name generation, and question answering---replicating previously proposed approaches while further improving the performance. 
We demonstrate the effectiveness of \approach{} for the aforementioned tasks using various LLMs. Our systematic evaluation compares different state machine configurations against baseline approaches without state machines. Results show that integrating well-designed state machines significantly improves the quality of LLM outputs, and is particularly beneficial for complex tasks with well-established human best practices but lacking data used for training LLMs.

\end{abstract}



\section{Introduction}

\noindent \textbf{Context.} 
Models capturing structural system behavior are fundamental elements of model-driven engineering (MDE). Behavioral models such as hierarchical state machines~\cite{HAREL1987231}, activity diagrams~\cite{jacobson2021unified}, and sequence diagrams~\cite{jacobson2021unified} provide well-defined structures to represent system behavior at various levels of abstraction. Within the context of MDE, these models facilitate the mapping of requirements to system design by leveraging human best practices and domain knowledge. Subsequently, these models enable the generation of lower-level artifacts. They also support the verification of system behavior, which is essential for safety-critical applications.

Large language models (LLMs) produce stochastic sequences of predictions based on input prompts~\cite{brown2020language-gpt3}. With the emergence of advanced LLMs such as GPT-4o~\cite{hurst2024gpt} and Qwen-2.5~\cite{yang2024qwen2}, they have become powerful tools for automating complex tasks across diverse domains using natural language. LLMs have shown remarkable capabilities in various applications, including code generation~\cite{jiang2024survey}, question answering~\cite{shailendra2024survey}, and planning~\cite{huang2024understanding,aghzal2025survey}. In MDE, recent efforts have also shown promising results in fully automating the generation of diverse types of models directly from natural language descriptions~\cite{di2025use}.

\noindent \textbf{Problem description.}
However, the stochastic nature of LLMs raises significant concerns around the potential hallucination in the generated outputs~\cite{huang2025survey}. Moreover, specific human best practice is rarely reflected in training data, which limits LLMs' performance in complex and domain-specific tasks. 
Approaches like retrieval augmented generation (RAG) provide a way to add up-to-date task context~\cite{lewis2020retrieval, asai2023self, jiang2023active, khattab2022demonstrate}, but they do not offer an approach to solve tasks involving complex workflows. 
Despite recent advancements in multi-step reasoning approaches such as Chain-of-Thought~\cite{wei2022chain}, Tree-of-Thought~\cite{yao2024tree}, and ReAct~\cite{yao2023react}, LLMs still face challenges in maintaining long-term consistency and performing effective multi-step planning for complex tasks. 

Recently, integrating structured workflows into the execution of LLM-based applications has been recognized as a promising solution to address these limitations. Multiple frameworks \cite{wu2024stateflow, liu2023smot} have been proposed to integrate LLMs with behavioral models to enhance task performance. However, the model used in these approaches still remains simple and is often coupled with implementation. 
More expressive behavior models, such as hierarchical state machines, which provide flexible and powerful mechanisms to decompose tasks into simpler sub-tasks, have not yet been fully leveraged. 
Moreover, best practices can often be modeled in multiple ways, which may significantly influence the performance of LLM systems, making it important for the framework to enable rapid experimentation.

\noindent \textbf{Objectives.}
This paper investigates the impact of model-driven processes on LLM task execution using behavioral models to represent human best practices. We explore the potential of these models to enhance the performance of LLMs for complex tasks, by proposing \approach{}: a framework for \texttt{S}tateful \texttt{H}ierarchical \texttt{E}xecution and \texttt{R}easoning for \texttt{P}rocess \texttt{A}utomation. \approach{} leverages hierarchical state machines to explicitly incorporate best practices into the task execution process. LLMs can be used within actions performed during state transitions. Additionally, \approach{} supports flexible task execution by enabling state transitions driven by logic and stochastic predictions,
while explicitly preserving execution histories and intermediate results for subsequent executions.

\noindent \textbf{Contribution.}
This paper makes the following contributions:
\begin{itemize}
    \item We propose \approach{}, a model-driven framework that integrates human best practices as hierarchical state machines with LLMs to enhance task performance.
    \item We propose a hybrid policy for state transition that combines rule-based and LLM-driven approaches while proposing a belief structure to store the execution history and intermediate results.
    \item We show the applicability of \approach{} to a variety of tasks, including code generation, class name generation, and question answering. We demonstrate how \approach{} can be used to replicate existing approaches while improving their performance by enhancing the state machine design.
    \item We evaluate the effectiveness of \approach{} on these three tasks using two LLM families: GPT-4o~\cite{hurst2024gpt} and Qwen-2.5~\cite{yang2024qwen2} as well as task-specific LLMs such as Qwen-2.5-Coder~\cite{hui2024qwen2} and Llama3.1-70B~\cite{grattafiori2024llama}.
\end{itemize}

\noindent \textbf{Added value.}
Compared to existing approaches combining LLMs with structured workflows, \approach{} is a more general, model-driven framework for tackling complex tasks with LLMs.
It showcases how current LLMs can benefit from MDE approaches by enabling task decomposition with modular representation. \approach{} decouples the state machine design from the action implementation, allowing changing state machine designs without updating the implementation. 

    
\section{Background}
\noindent
\textbf{LLMs.} Large language models (LLMs) have gained significant attention for many tasks based on natural language~\cite{shailendra2024survey,jiang2024survey,di2025use}. LLMs use deep neural networks, typically with transformer architecture~\cite{vaswani2017attention}, to estimate a probability distribution of text sequences. Current state-of-the-art generative LLMs are primarily based on the decoder-only architecture, which is derived from the original transformer architecture. In general, an \llm{} $\texttt{llm}$ can be seen as a mapping between sequences of input and output tokens: $\texttt{llm:} s_{in} \rightarrow s_{out}$. More specifically, given a sequence of tokens (called \emph{prompt}) $s_{in}=\{s_1, s_2,...,s_{k - 1}\}$, LLMs estimate the conditional probability of the next token $P(s_k|s_1, ..., s_{k - 1})$. Then \llm{}s continue to generate the next tokens auto-regressively until a special \emph{stop} token is generated or a maximum length is reached to output the final sequence $s_{out}$. These \llm{}s are typically first pre-trained on a large corpus of text data using unsupervised learning, followed by fine-tuning to follow input instructions~\cite{zhang2023instruction,ouyang2022training}.

\noindent
\textbf{LLM frameworks.}
Recent studies indicate that augmenting \llm{}s with external tools can enhance performance across various tasks~\cite{qin2023toolllm}. Such integration aims to separate task planning and decomposition with LLMs from task solving with other tools ~\cite{yao2023react} to address specific problems more effectively.

However, the performance of \llm{}s may be limited in tasks requiring domain-specific knowledge not already available in their training data. For instance, within the context of software modeling, \llm{}s struggle to generate domain models that adhere to established best practices and modeling patterns~\cite{chen2023automated}. To overcome this limitation, recent \llm{} frameworks support the construction of structured workflows to guide the behavior of \llm{}s~\cite{langgraph,wu2024stateflow}. In these frameworks, workflows are represented as graphs, with nodes corresponding to different tasks which can be potentially executed by an \llm{}, and edges denoting task progression. \llm{}s can also assist in making branching decisions at nodes within these workflows.

Furthermore, \llm{}s can be equipped with external memory to store and retrieve contextual information as needed~\cite{langgraph,packer2023memgpt}. This capability addresses their limitations in handling tasks across multiple steps. Such memory implementations often utilize key-value stores~\cite{packer2023memgpt}, enabling the retrieval of relevant information at various task stages.

In this paper, we aim to combine these aspects for LLMs by leveraging a state-machine-driven approach. Each instance of \approach{} incorporates a hierarchical state machine to structure workflows, where actions within transitions may use external tools. During state machine execution, a \emph{belief} is maintained as memory to retain and access pertinent information.

\noindent
\textbf{State machines.}
A state machine (SM)~\cite{mealy1955method} is a model used to describe the behavior of a system that can be in a finite number of states. SMs provide a visual representation to help with the development and maintenance of a system. Moreover, they can be used to enable automated code generation~\cite{domi2012systematic,van2017complete} and facilitate verification of system behaviors~\cite{alur1998model}.

Formally, an SM is a tuple $(\set{S}, \set{T}, \set{A}, \set{G}, entry, exit, \delta, s_0, \set{F})$ where $\set{S}$ is a set of states, $\set{T}$ is a set of event triggers for transitions, $\set{A}$ is the set of actions that can be performed during transitions or in states, $\set{G}$ is the set of guard conditions for transitions, $entry: \set{S} \to \set{A}$ and $exit: \set{S} \to \set{A}$ map states to entry and exit actions, respectively. $\delta: \set{S}\times \set{T} \times \set{G} \to \set{A} \times \set{S}$ is the state transition function, $s_0 \in \set{S}$ is the initial state, and $F \subseteq \set{S}$ is the set of final states. 

A hierarchical SM is an extension of SMs~\cite{HAREL1987231} by organizing states hierarchically. When a state is activated, all its parent states are also considered active, and transitions defined for a parent state apply to its child states as well. This hierarchy improves modularity and abstraction, reducing redundancy in defining transitions between related states. Formally, a hierarchical SM adds a hierarchical function $h: \set{S} \to \set{S}$ that maps states to their parent states. The state transition function $\delta$ is extended to include the parent state in the transition tuple. 

There exist many definitions of the textual languages for SMs, such as PlantUML~\cite{plantuml} and Umple~\cite{lethbridge2021umple}. While \approach{} uses JSON to define the SM, it can be easily adapted to other concrete syntax by using a parser. 

\section{Combining State Machines with LLMs}

This section introduces how \approach{} integrates SMs with \llm{} execution. We first provide an overview of the \approach{} framework, followed by a detailed description of the architecture with an example in question answering. 


\subsection{Overview}

\begin{figure}[t]
    \centering
    \includegraphics[width=0.85\linewidth]{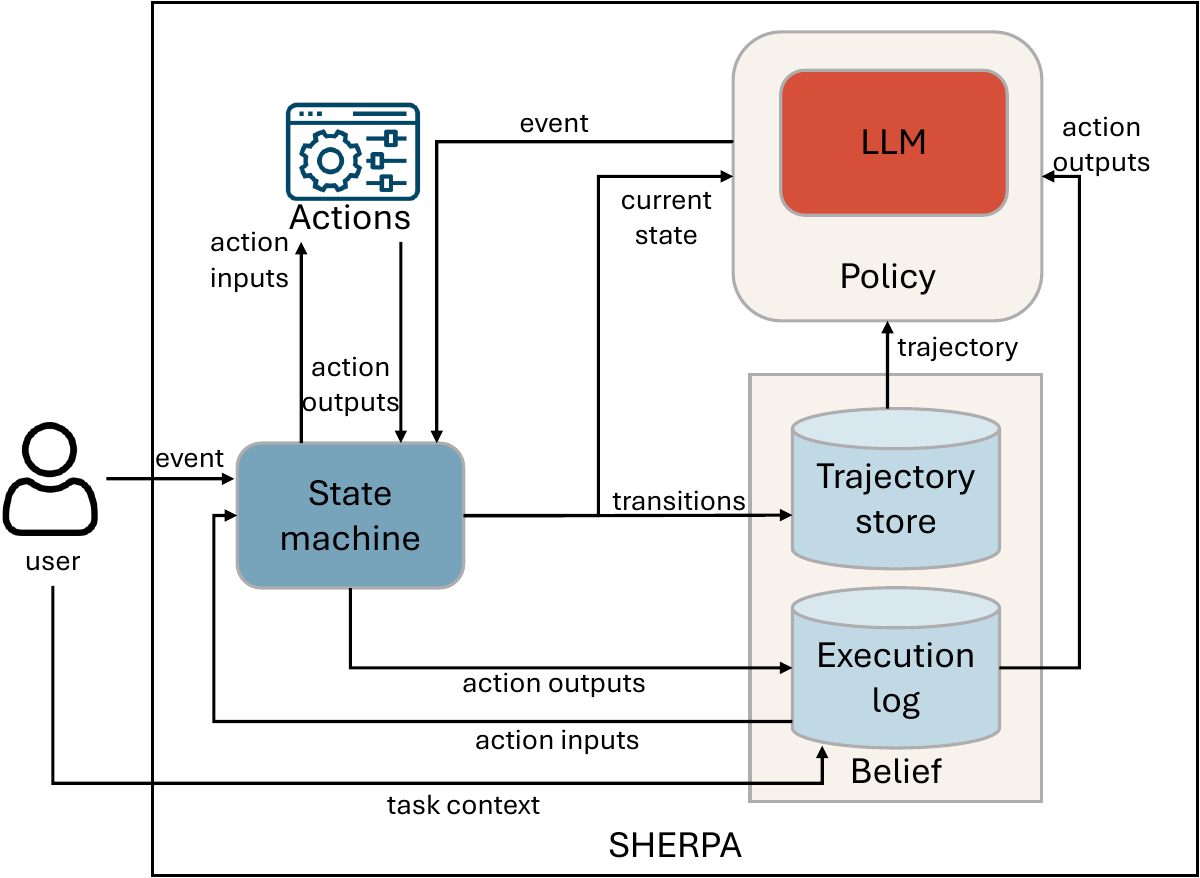}
    \caption{Architecture for \approach{}}
    \label{fig:architecture}
\end{figure}

\autoref{fig:architecture} provides an overview of the architecture. We define each instance of \approach{} as an \emph{agent} that is capable of executing a specific task. Each agent is associated with a \emph{state machine}. A user, either a human or another agent, interacts with the agent by sending an \textit{event} to its SM. These events can include messages, such as a question, which are recorded as \emph{task context} in the agent's \emph{belief}.
Then the SM performs state transitions based on the received event and may execute predefined \textit{actions} associated with that transition. Following the transition, the SM updates the agent's \textit{belief} with the result of the action and the agent's new state. The belief contains the \textit{trajectory store} keeping track of the taken \textit{transitions} in the SM, and an \textit{execution log} containing (1) executed actions including their \textit{outputs} and (2) a key-value store containing information relevant to the subsequent states as action \textit{inputs}.

This updated belief, along with available transitions in the \textit{current state}, is then provided to the \textit{policy}, which uses an \llm{} or a predefined rule to determine the optimal next step. The selected event is subsequently sent back to the SM to initiate further transitions. This cycle continues until the SM either reaches an end state (terminating the process) or all transitions require external input (e.g., waiting for further user input). Next, we provide a detailed introduction to each component.


\subsection{Architecture}
\begin{figure}[t]
    \centering
    \includegraphics[width=0.85\linewidth]{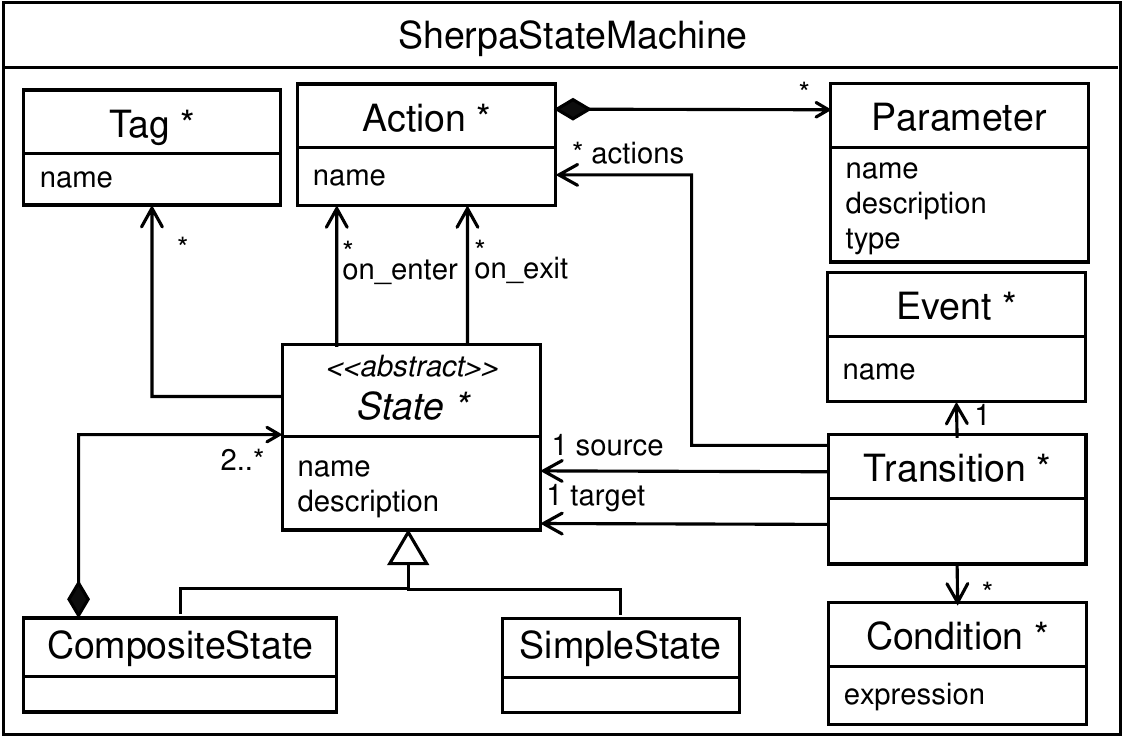}
    \caption{The metamodel for state machines in \approach{}}
    \label{fig:domain_model}
\end{figure}
\paragraph{State machine}
The SM in \approach{} systematically controls the LLM behavior by defining explicit states and transitions inspired by human best practices for the targeted task. Note that the state machine is a type of \textit{data} in \approach{}, which supports dynamically altering the structure of the state machine without changing other components for an agent.
\autoref{fig:domain_model} shows the metamodel of the SM used in \approach{}.

The core components of the SM are its \textbf{states}. Although standard UML SMs include various types of states, such as parallel regions and history states, hierarchical states are particularly useful for decomposing complex tasks into simpler, manageable sub-tasks. Such decomposition aligns closely with the primary purpose of integrating human best practices into LLMs. Therefore, \approach{} employs two types of states: atomic \textbf{simple states} and hierarchical \textbf{composite states}. A composite state contains multiple sub-states. Each state includes a \textit{name} and a textual \textit{description}, which provide meaningful representations of the current state. These attributes are essential for the policy to evaluate and select potential transitions, especially when an LLM is used within the policy. Additionally, states may have optional \textbf{tags} denoting special behaviors. Currently, the supported tags include \textit{start} and \textit{end}, indicating the initial and final states, respectively.

The SM also incorporates a set of \textbf{transitions} defining its operational flow. Each transition explicitly connects a \emph{source} state to a \emph{target} state and specifies an \textbf{event} that triggers this transition. The triggering event may originate from a user message or be internally generated by the agent’s policy. Transitions may also include optional guard conditions to provide finer-grained control over state transitions. Such guard conditions are represented using the \textbf{condition} class in the form of either a simple logical \textit{expression} evaluated against the agent's belief or the \emph{name} of an action with more sophisticated logic, such as invoking another \llm{}.

The behavior of the agent during state transitions is defined by \textbf{actions}. Typically, actions perform sub-tasks such as calling an LLM, retrieving relevant data from a database, or calling external tools for various types of specialized computation. Each action is characterized by a \emph{name} and a set of \textbf{parameters}, which are defined by their \textit{names}, \textit{types}, and \textit{descriptions}, including their expected data type and guidance for how they should be provided. Each parameter in an action can be one of the following two \textit{types}, depending on how it should be provided: (1) \emph{external parameters} are provided externally when the action is invoked, such as user input or provided by the policy through an LLM; (2) \emph{internal parameters} are provided internally through the agent's belief during the execution of the action.
Actions can be integrated into the SM in multiple ways: they may be attached directly to transitions, triggering execution upon transition activation, or associated with states, executing when \emph{entering} or \emph{exiting} a particular state.

\paragraph{Policy}
A \emph{policy} in \approach{} determines the next transition for an agent by considering its current state and belief. The term policy is adapted from reinforcement learning literature, given its analogous role in decision-making \cite{sutton1998reinforcement}. Formally, let $s \in \set{S}$ be the current state, $\set{T}_s$ be the set of available transitions at state $s$, $\set{E}$ be the set of possible events that can trigger transitions, and $B$ be the agent’s belief. A policy is defined as a mapping $\pi(s, \set{T}_s, B) \to e$, where $e \in \set{E}$ is the selected event along with any data required by the transitions.

Policies can be implemented using various approaches, including rule-based systems, a supervised machine learning model, reinforcement learning, or \llm{}s. In the current version of \approach{}, to demonstrate the concept, we introduce two primary policy types that can be combined flexibly to guide an agent's behavior. Notably, \approach{} can be easily extended to incorporate additional policies provided that they conform to the predefined policy interface.

\textbf{Rule-based policy.} 
A rule-based policy uses predefined mapping rules to determine the next transition based on the agent's current state and belief. These rules typically reference the current state and specific values stored within the belief of an agent. For example, a rule may specify that whenever a generated domain model includes an abstract class without subclasses, the policy should select a transition leading back to a state responsible for regenerating the domain model.

\begin{figure}[tb]
    \centering
    \includegraphics[width=0.85\linewidth]{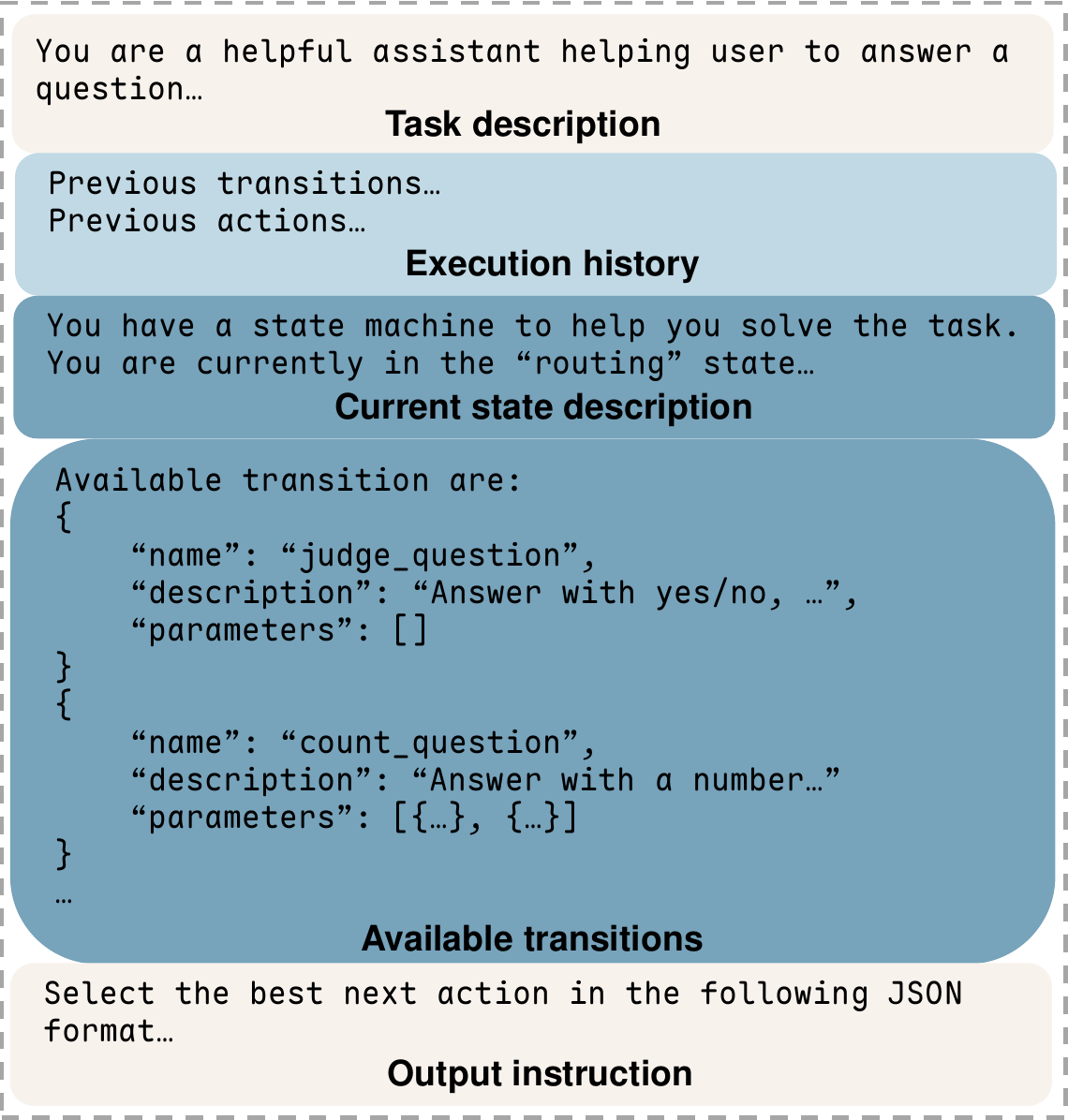}
    \caption{Prompt for the LLM-based policy; the execution history including the trajectory and actions comes from the belief while the current state description and available transitions are provided by the SM.}
    \label{fig:policy_prompt}
\end{figure}

\textbf{LLM-based policy.} 
Inspired by the ReAct framework \cite{yao2023react}, the LLM-based policy utilizes an \llm{} to determine the next state transition. \autoref{fig:policy_prompt} illustrates an example prompt used for selecting the next transition within the context of question answering. Upon policy initialization, a \emph{task description} defining the overall task, along with an \emph{output instruction} specifying the expected output, is provided. During execution, when the LLM-based policy is invoked, the relevant \emph{execution history}, including the trajectory and previous actions, is extracted from the \emph{belief}. To prevent exceeding the maximum token limit of \llm{}s, \approach{} uses a sliding-window mechanism to extract the most recent execution history that uses up to a fixed amount of tokens. Additionally, the SM provides the \emph{current state}, including its description and available transitions. If any actions are associated with these transitions, or with exiting the current state or entering the subsequent state, the \llm{} is also prompted to generate the necessary \emph{external input parameters} for these actions. Based on this information, the \llm{} is then prompted to select the most appropriate next transition.

Beyond these two policy types, \approach{} incorporates a \emph{fast-forward} mechanism for efficiency. In situations where only one transition is available, due to either the SM’s structure or the evaluation of guard conditions, the fast-forward mechanism automatically selects this transition, skipping potentially computationally intensive policy evaluations.

\paragraph{Belief}
The \emph{belief} of an agent in \approach{} is a structured data store to maintain information relevant to the current task. Compared to typical memory in LLM-based agents \cite{packer2023memgpt}, which usually consists solely of a key-value store, the belief structure in \approach{} additionally includes a \emph{trajectory store} and an \emph{execution log}. The trajectory store maintains a sequence of state transitions that the agent has traversed, thereby providing essential context for decision-making. In parallel, the execution log records the history of actions performed by the agent, enabling tracking and retrospective analysis of behavior. The execution log also contains a key-value store preserving task-specific information that can be dynamically saved during action execution. Information in the belief is accessible during both transition selection by the policy and action execution.

\begin{figure}[tb]
    \centering
    \includegraphics[width=0.8\linewidth]{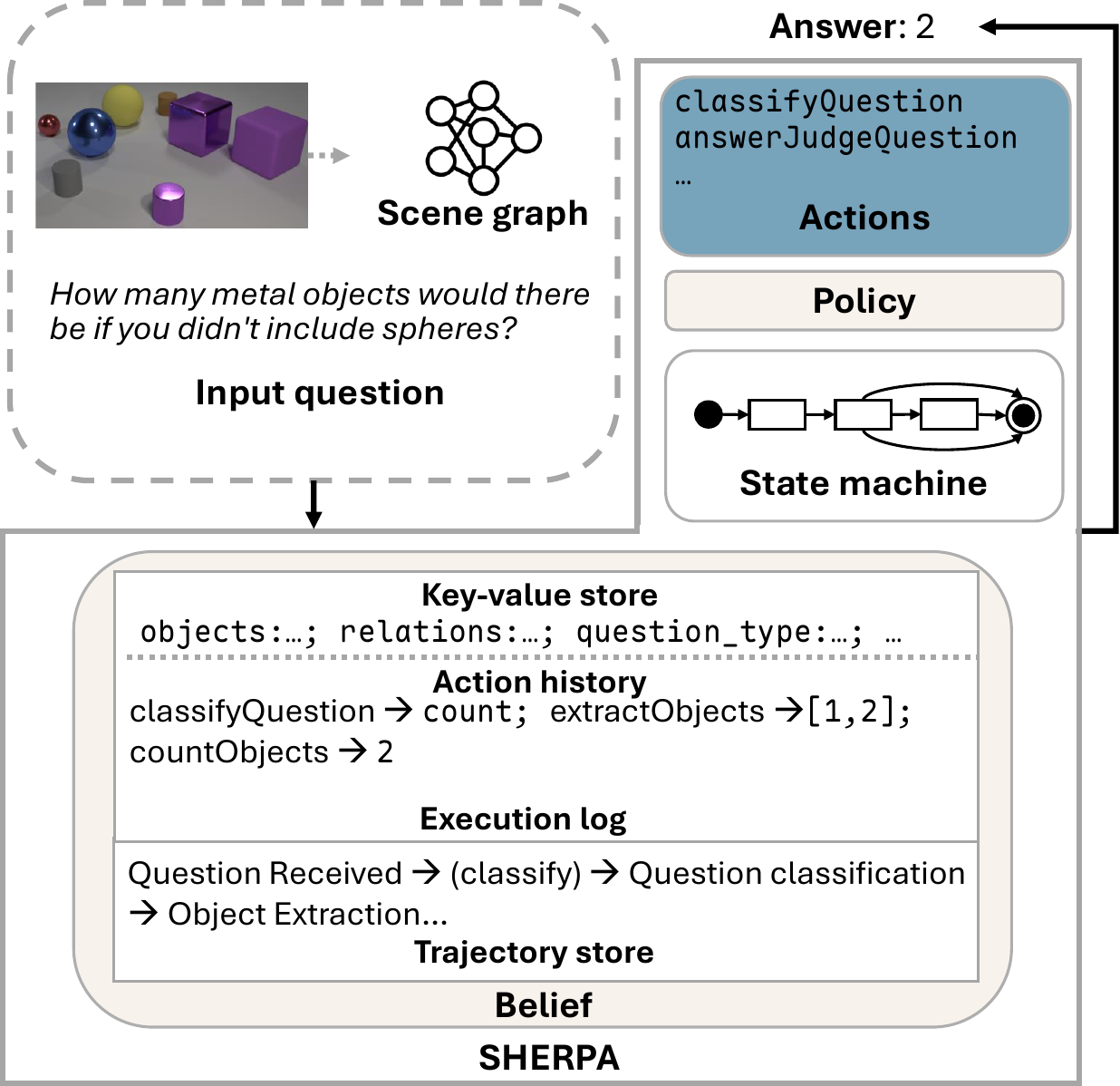}
    \caption{Example of \approach{} for the scene graph-based question answering}
    \label{fig:example}
\end{figure}

\subsection{Example}
\begin{figure*}
    \centering
    \includegraphics[width=\linewidth]{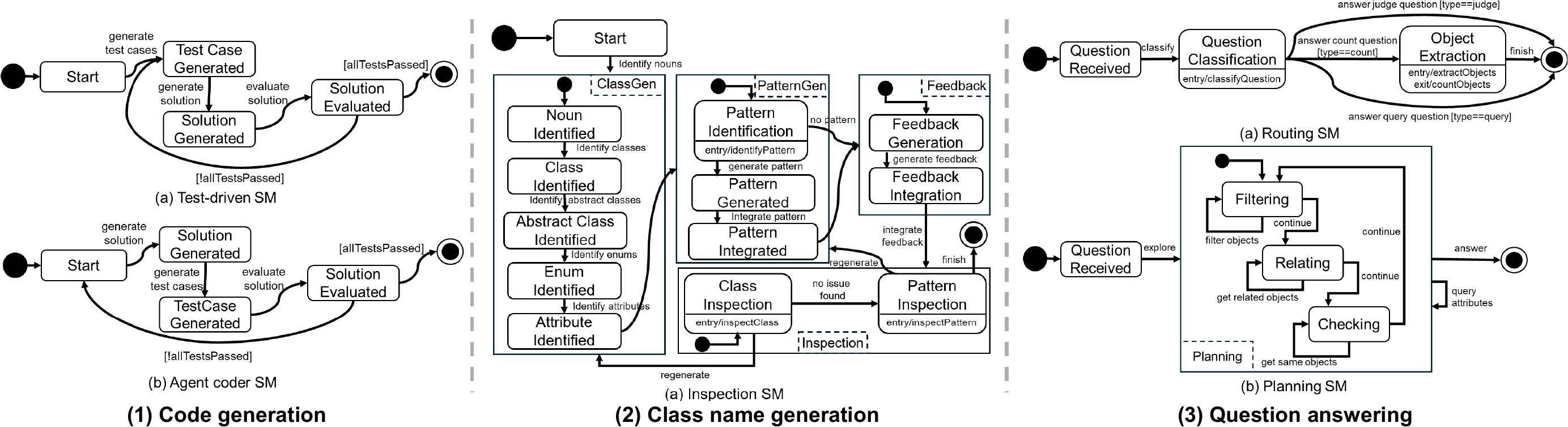}
    \caption{SM design for the use cases; states for outputs and actions on the transitions are omitted for conciseness} 
    \label{fig:use_case}
\end{figure*}

\autoref{fig:example} illustrates the integration of a routing SM (see \autoref{fig:use_case} (3.a)) for question answering from a scene graph. Specifically, the LLM is prompted with the question: \emph{"How many metal objects would there be if you didn't include spheres?"}. Note that the image is first converted into a textual scene graph, and the LLM's task is to answer the given question based on this graph. Initial experiments indicate that LLMs frequently produce incorrect answers to counting questions. We hypothesize that this arises because the LLM simultaneously performs multiple complex subtasks: (1) interpreting the question, (2) extracting relevant objects from the scene graph, and (3) counting these extracted objects.

To address this challenge, we propose employing an SM to explicitly decompose the question-answering process into distinct subtasks based on the type of question. Specifically, relevant contextual information, such as the scene graph and question, is initially stored in the \emph{belief}. Then, the LLM is prompted to determine the question type (\emph{classifyQuestion} action within the \emph{QuestionClassification} state). If identified as a counting question, the LLM is subsequently prompted to extract relevant objects from the scene graph (\emph{extractObjects} action upon entering the \emph{ObjectExtraction} state). After extraction, the number of objects is counted deterministically (\emph{countObjects} action upon exiting the \emph{ObjectExtraction} state). Throughout this procedure, \approach{} maintains a \emph{trajectory store} to track state transitions, and an \emph{execution log} that records the sequence of executed actions along with their corresponding inputs and outputs. Finally, the output of \approach{} is defined as the result of the final executed action. 


\section{Use Cases}
\label{sec:use_case}

To demonstrate and assess the effectiveness of \approach{}, we examine three distinct tasks that can be automated using LLMs: \emph{code generation}, \emph{class name generation}, and \emph{question-answering}. We specifically select these tasks due to their differences in data availability and human best practices. Code generation is widely recognized as a popular task for evaluating the capabilities of LLMs, supported by numerous available datasets~\cite{hou2024large}. Additionally, established human best practices, such as test-driven development, are already popular in software engineering, providing effective frameworks for addressing this task. In contrast, class name generation represents a typical modeling task. Although this task has well-established human best practices, it currently has significantly fewer available datasets. Finally, question-answering is a well-known task for LLM evaluation with many datasets \cite{yue2025survey}; however, unlike the previous two tasks, it lacks clearly defined human best practices for systematically solving this task generally.

\subsection{Code generation}

\noindent\textbf{Task description.} 
The code generation task 
involves generating code based on natural language descriptions. LLMs recently have emerged as promising tools for this task due to their strong code generation ability~\cite{hou2024large}.

\noindent\textbf{State machines.} 
To systematically address the code generation problem using \approach{}, we implement two SMs (see \autoref{fig:use_case} (1)) based on how humans use tests to write code.

In the \emph{test-driven} SM, we first ask the LLM to generate test cases based on the function description,
after which the LLM is prompted to generate the solution function. A function is returned if it passes all the tests. If not, the process returns to the \textit{Test Case Generated} state to retry. If no function passes all the tests after a certain budget is reached, the one that passes the most tests is returned.

The \emph{agent coder} SM is based on the AgentCoder approach~\cite{huang2023agentcoder}, with a single LLM performing both roles: programmer and test designer. The process begins with the LLM generating the function. Once the function is produced, the LLM then generates test cases. If the function does not pass all tests, the process retries from the \textit{Start} state, generating a new candidate function and tests. If no function passes all the tests, the one that passes the most tests is returned.

The \emph{agent coder} SM shows the applicability of \approach{} by replicating a previous approach. The \emph{test-driven} SM can be seen as a refinement of the \emph{agent coder} SM, where the repeated generation of test cases is avoided. 

\subsection{Class name generation}

\noindent\textbf{Task description.} 
Domain model generation is an important MDE task where LLMs are increasingly popular~\cite{di2025use}. This task involves generating a domain model based on a problem description. However, the evaluation of generated domain models is typically labor-intensive. Thus, we focus on generating class names in domain models to simplify this process. 


\noindent\textbf{State machines.} 
To systematically address this generation problem using \approach{}, we implement two SMs, one of which is shown in \autoref{fig:use_case} (2). The main flow in the SM is motivated by the multi-step iterative generation approach (MIG)~\cite{yang2024multi}, which includes the following components for class names: 

(1) \emph{Classes and attributes generation.} 
In this composite state, the LLM is asked to generate all potential classes, including regular classes, abstract classes, and enumerations. The process begins by extracting relevant nouns from the problem description, which serve as candidate class names, attribute names, or enumeration literals. Subsequently, the approach categorizes these candidates into regular, abstract, and enumeration classes in a cascading manner. The final step involves assigning attributes to each class based on previous noun analysis and previously identified class types.  

(2) \emph{Pattern generation.} 
Patterns are an integral part of domain modeling. In this composite state, the approach performs \emph{player-role} pattern identification and makes transitions based on the result using an LLM. If no designated pattern is detected, the SM transitions to the feedback generation state. Otherwise, it proceeds to generate the identified pattern and integrate the pattern into the model to align with modeling practice.  

(3) \emph{Generate feedback.} 
This composite state enhances the partial domain model through LLM self-reflection. The approach first generates feedback about the model, then iteratively improves the partial domain model based on this feedback. Even though the focus is on class name generation, attributes and patterns are identified as they also help to identify classes.

The above processes are largely \emph{linear} as there is no mechanism for the generation process to retry if some major mistakes are detected. To address this issue, we further introduce the \emph{inspect model} composite state, providing transitions to return to the \emph{identify classes and attributes} state or \emph{identify pattern} state. The LLM is asked to examine the classes and determine if the process needs to go back to a previous state.

\subsection{Question answering} 
\noindent\textbf{Task description.}
Question answering is a classic natural language processing task. Due to LLMs' impressive natural language understanding capabilities, they are increasingly used for this task~\cite{yue2025survey}. In this paper, we specifically focus on \emph{scene graph-based} question answering, where the input context is provided as a scene graph. We categorize the questions into three types: (1) \emph{judging} questions, which require determining whether a statement is true or false; (2) \emph{counting} questions, which ask how many objects within the scene graph satisfy a particular condition; and (3) \emph{querying} questions, which request specific attributes of objects within the scene graph. Due to the lack of established human best practices, we propose three SM approaches inspired by popular LLM-based methods~\cite{varangot2025doing,yao2023react} (two of which are shown in \autoref{fig:use_case} (3)).

\noindent\textbf{State machines.}
The \emph{routing} SM follows the approach demonstrated in the example in \autoref{fig:example}, representing a typical question-answering strategy using LLMs~\cite{varangot2025doing}. The core idea is to first classify the question type and then address it using methods tailored specifically for each type. For counting questions, the LLM is prompted to extract objects matching the question criteria, after which counting is performed deterministically. In contrast, for judging and querying questions, the LLM directly generates the answers.

We implement the widely adopted ReAct approach~\cite{yao2023react} for leveraging LLM reasoning with \approach{} as \emph{ReAct} SM (not shown in the figure for conciseness). 
Here, the primary role of the LLM is to perform planning by sequentially calling predefined operations on the scene graph to gather relevant information. Thus, the SM consists of one state with self-transitions for these operations. The available operations are defined as follows: (1) \emph{filter}, to extract objects from the scene graph based on attribute values; (2) \emph{relation}, to identify objects having specific relations to another object; (3) \emph{checking}, to determine objects sharing a particular attribute with another object; and (4) \emph{query}, to retrieve an attribute of a given object. Finally, there is another transition to the end state. 

However, we hypothesize that providing too many action choices might overwhelm the LLM. Therefore, we propose the \emph{planning} SM, a more structured version of the \emph{ReAct} SM. This approach provides a predefined action sequence as heuristic based on observations from sampled questions. Specifically, the LLM first \emph{filters} the scene graph based on attributes, then retrieves objects either \emph{related} to those filtered objects or by \emph{checking} objects sharing the same attributes. At any step, the LLM can \emph{query} attributes of any object in the scene graph.

\section{Evaluation}
Given a use case, there are multiple ways to implement best practices into an SM to guide the behavior of LLMs. Furthermore, while \approach{} can be used to implement existing LLM approaches, it is not trivial to determine the best design for a given use case.  
In this section, we evaluate popular approaches and their enhanced version implemented with \approach{} using the defined use cases. In particular, we target to address the following research questions:
\begin{itemize} 
    \item \textbf{RQ1} How does integrating state machines in \approach{} impact the performance of LLMs with different sizes compared with directly using LLMs?
    \item \textbf{RQ2} How does the state machine design and configuration affect the performance and cost of \approach{}?
\end{itemize}

\subsection{Evaluation setup}
\paragraph{LLMs}
We evaluate \approach{} using four popular LLMs for all use cases as discussed in \autoref{sec:use_case}. Additionally, we choose one dedicated LLM for each use case that was designed to solve the task targeted by that specific use case. 

\noindent\textbf{Common LLMs.}
We evaluate our approach using two of the most recent LLMs from OpenAI: GPT-4o Mini~\cite{gpt-4o-mini} and GPT-4o~\cite{hurst2024gpt}. These models demonstrate improved capability and performance on complex tasks compared to their predecessors, while reducing inference costs. Nevertheless, their closed-source nature raises concerns regarding data privacy, security, and reproducibility. Thus, we also perform experiments with two state-of-the-art open-source models: Qwen2.5 7B~\cite{yang2024qwen2}, one of the most powerful LLMs deployable on consumer-grade GPUs, and Qwen2.5 72B, one of the most capable configurations within the Qwen LLM family~\cite{yang2024qwen2}. Both LLMs exhibit better performance over similarly scaled open-source alternatives across a wide range of evaluation tasks.

\noindent\textbf{Task-specific LLMs.}
We also select specialized LLMs tailored specifically to each use case. For the \textbf{code generation} task, we choose Qwen2.5 Coder (32B)~\cite{hui2024qwen2}, an open-source LLM designed for code generation, achieving top performance across multiple coding benchmarks. For the \textbf{class name generation} and \textbf{question answering} tasks, primarily challenging natural language-related capabilities, we evaluate Llama3.1 70B~\cite{grattafiori2024llama}, another popular and powerful open-source LLM comparable to Qwen2.5 72b but outside the GPT and Qwen families.

\paragraph{Benchmarks} 
We evaluate the effectiveness of different approaches with the following benchmarks. 

\noindent \textbf{Code generation.}
To evaluate how \approach{} enhances the code generation process, we use the \textbf{HumanEval} benchmark~\cite{chen2021evaluating}. The benchmark includes 164  programming problems and is widely adopted in LLM research to assess model performance in code generation~\cite{jiang2024survey}. In this benchmark, the LLM is provided with a natural language description and the header of a Python function and is tasked with generating the function body. 


\noindent \textbf{Class name generation.}
For class name generation, we use the automated domain modeling dataset proposed by Chen et al. \cite{chen2023automated} (\textbf{Modeling}). This dataset contains eight domain models with a total of 135 classes, spanning diverse domains and varying complexity levels. We specifically use classes from these domain models for evaluation.  


\noindent \textbf{Question answering.}
For the scene-based question answering task, we use human-curated questions from the \textbf{Clevr} dataset \cite{johnson2017clevr,johnson2017inferring}. Compared to template-generated questions, these human-curated questions offer greater linguistic diversity and complexity. We manually verify each question and exclude any questions that require visual context beyond the scene graph (e.g., object reflections). In total, we randomly sample 100 questions from the dataset, including 33 judgment-based, 33 counting-based, and 34 querying-based questions.

\paragraph{Metrics} 
We select metrics to evaluate the quality of results specifically for each use case.

\noindent \textbf{Code generation.} 
Each generated function is evaluated using the test suite from the HumanEval benchmark~\cite{chen2021evaluating}. We use the \pass{} metric, measuring the proportion of problems solved by passing the test suite within $1$ attempt. 

\noindent \textbf{Class name generation.} 
Evaluating class names can be challenging due to variations in synonyms and naming conventions. Relying solely on exact matches typically significantly underestimates the true generation quality. To overcome this limitation, we adopt an embedding-based evaluation method~\cite{chen2024embedding}, which has been shown to correlate well with human judgments.
This approach measures the cosine similarity between generated and reference labels to create overall scores (precision (\precision{}), recall (\recall{}), and $F_1$-score (\fscore{})) for the set of generated classes. 

\noindent \textbf{Question answering.}
For Clevr, we measure answer accuracy (\accuracy{}). As the answers belong to a finite set of categories (e.g., numbers, 
attribute values, etc.), we use \textit{exact match} accuracy between the generated and ground truth answers.

Furthermore, the iterative nature of the SMs can potentially lead to increased cost for LLM invocations, especially when using an LLM API or when the LLM is deployed on a cloud service. To evaluate this aspect, we also measure the \emph{average number of LLM calls} required for each use case.

\paragraph{Implementation details}
We set the number of maximum state transitions to 10 for each use case. To overcome the variability of the LLM's output, we choose a 0.01 generation temperature and measure the average performance over 3 runs. The repository for the evaluation is available online \cite{artifact_models2025}.

\subsection{RQ1: Comparison with direct methods}
\paragraph{Rationale and setup}
When considering building domain-specific SMs to control LLM behavior for a particular task, it is crucial to understand the situations under which such an approach can be beneficial. This research question addresses this aspect by comparing the performance of SMs powered by \approach{} against the \direct{} approach, which relies solely on prompting techniques. We investigate two sub-questions:
\begin{itemize} 
    \item \textbf{RQ1.1:} How do SMs in \approach{} influence the quality of outputs from LLMs compared to \direct{}? 
    \item \textbf{RQ1.2:} How does the impact of SMs vary across different LLM sizes within the same LLM family? 
\end{itemize}

For each use case, we select appropriate prompting techniques for the \direct{} method. Specifically, for the code generation task, we choose zero-shot prompting using a widely adopted prompt that has demonstrated effectiveness across various GPT models~\cite{humaneval_prompt}. For the class name generation task, we utilize one-shot prompting, providing a single illustrative example from a domain not included in the testing set, demonstrating the nature of the task and the expected format of the output. Lastly, for the question answering task, we adopt chain-of-thought prompting~\cite{wei2022chain}, as answering questions based on scene graphs typically requires step-by-step reasoning.

For the SM in \approach{}, we use the first SM illustrated in \autoref{fig:use_case} for each use case (specifically, the test-driven, inspection, and routing SMs). Note that different SM designs can significantly influence the quality of the final output (see RQ2). For RQ1, our goal is to understand the general impact of using an SM. Therefore, we selected SMs representing straightforward designs that can be easily constructed and implemented by software engineers.

\begin{table*}[tb]
    \centering
    \caption{Performance comparison between direct and SM-based methods.}
    \scalebox{0.8}{
    \begin{tabular}{|c||c|c||ccc|ccc||c|c|}
        \hline
        & \multicolumn{2}{c||}{HumanEval} & \multicolumn{6}{c||}{Modeling} & \multicolumn{2}{c|}{Clevr}\\
        \hline
        & \direct{} & \approach{} &  \multicolumn{3}{c|}{\direct{}} & \multicolumn{3}{c||}{\approach{}} & \direct{} & \approach{} \\
        \hline
        & \pass{} & \pass{} & \precision{} & \recall{} & \fscore{} & \precision{} & \recall{} & \fscore{} & \accuracy{} & \accuracy{} \\
        \hline
        GPT-4o Mini & 85.37 & \underline{88.82} & 88.89 & 63.20 & 73.36 & 85.22 & 70.77 & \underline{76.77} & 86.00 & \underline{86.67}  \\
        \hline
        GPT-4o & \underline{91.26} & 90.24 & 92.98 & 55.24 & 68.90 & 92.82 & 58.65 & \underline{71.16} & \underline{\textbf{90.67}} & 88.33  \\
        \hline
        Qwen2.5 7B & 83.13 & \underline{85.37} & 81.15 & 47.52 & 57.38 & 89.75 & 62.06 & \underline{71.57} & 70.00 & \underline{75.33}  \\
        \hline
        Qwen2.5 72B & 84.76 & \underline{90.65} & 90.41 & 56.78 & 68.98 & 89.35 & 73.71 & \underline{\textbf{80.07}} & \underline{86.67} & \underline{86.67}  \\
        \hline
        Task-Specific LLM & 89.84 & \underline{\textbf{91.26}} & 88.76 & 64.07 & 73.85 & 88.51 & 67.68 & \underline{75.80} & 83.00 & \underline{84.67}  \\
        \hline
        
    \end{tabular}
        }
    \label{tab:rq1}
\end{table*}

\paragraph{Impact of SM integration}
\autoref{tab:rq1} presents the performance comparison between \approach{} and \direct{} across all use cases and LLMs. For each use case-LLM pair, we underline the better-performing method and highlight the best overall performance for each use case in bold. In general, empowering LLMs with SMs in \approach{} improves performance compared to the \direct{} approach in 12 out of 15 cases, while being comparable in the remaining three. \approach{} also achieves the highest performance in two out of three use cases, slightly underperforming GPT-4o in the Clevr dataset.

Analyzing the impact for each task individually, \approach{} outperforms \direct{} in code generation for four out of five LLMs, showing an average improvement of 3.25 percentage points (i.e., \approach{} - \direct{} = 3.25). The overall improvement in LLMs other than GPT-4o is primarily due to the explicit test-generation step included in the SM, which validates generated code against these tests.

For class name generation, integrating the SM notably improves recall but occasionally leads to decreases in precision. The iterative refinement in the SM results in more class names being discovered, but also occasionally introduces irrelevant class names. Nonetheless, \approach{} consistently achieves higher overall \fscore{} for all LLMs, with an average improvement of 6.58 percentage points. This substantial gain can be attributed to the SM's alignment with existing human best practices.

In the question-answering task, \approach{} achieves higher accuracy in three out of five LLMs, with an average improvement of 2.56 percentage points. Interestingly, it does not bring benefit with the two largest LLMs, likely because these models already effectively manage CLEVR’s question-answering tasks for different question types. Smaller LLMs, however, clearly benefit from the routing logic provided by the SM, which helps them address different aspects of each question type.

\begin{tcolorbox}[boxsep=-1mm]
    \textbf{RQ1.1.}
    Integrating SMs within \approach{} generally enhances LLM performance compared to the \direct{} method, improving the performance in 12 out of 15 cases. The most significant benefit occurs in the class name generation task, with an average improvement of 6.58 percentage points, which can be attributed to the fact that there exists an established best practice for this task.
\end{tcolorbox}

\paragraph{Impact of LLM size}
Examining the impact of LLM size on performance, we find that integrating SMs provides more consistent benefits for smaller LLMs. Within both the Qwen and GPT families, \approach{} consistently outperforms \direct{} for smaller LLMs, while it improves performance in only three out of six cases for larger variants. 
We hypothesize that larger LLMs possess sufficient capabilities to handle the tasks effectively without integrating SMs.
Indeed, larger LLMs consistently outperform their smaller counterparts across all tasks, except in one of the class name generation cases, where GPT-4o Mini unexpectedly outperforms GPT-4o. We suspect that the benefits observed in smaller LLMs may also occur in larger LLMs when applied to more complex tasks.

During early experiments, we iteratively developed the SM with smaller LLMs to reduce inference costs. As a result, the final SM design may favor smaller LLMs and limit effectiveness with larger variants, suggesting that SMs should be tailored to the target LLM to maximize their benefits.

\begin{tcolorbox}[boxsep=-1mm]
    \textbf{RQ1.2.}
    The impact of SMs on LLM performance is more pronounced in smaller models. Specifically, \approach{} consistently outperforms \direct{} for smaller LLMs, while it may damage the performance for larger LLMs. This observation indicates that the SM design should be tailored to the specific LLM to maximize effectiveness. 
\end{tcolorbox}

\subsection{RQ2: Impact of SM design}
\paragraph{Rationale and setup}
RQ1 demonstrated that integrating SMs with \approach{} generally improves the performance. However, the SM itself may also influence the effectiveness. Moreover, the multi-step nature of SMs increases the number of LLM calls and thus impacts the overall cost of LLM executions. In \approach{}, SM design is decoupled from action implementation, enabling updating the SM without modifying the underlying implementation to reach cost targets. In this RQ, we analyze how different SMs affect both performance and cost. Specifically, we address the following sub-questions:

\begin{itemize} 
    \item \textbf{RQ2.1:} How do different SM configurations in \approach{} influence the quality of LLM-generated outputs? 
    \item \textbf{RQ2.2:} How do different SM configurations in \approach{} impact the number of LLM invocations? 
\end{itemize}

We use SMs for the use cases described in \autoref{sec:use_case} to answer these questions. For the code generation task, we compare the \emph{test-driven} SM, which attempts to reduce LLM calls by avoiding repeatedly generating new test cases when a generated function fails, to the \emph{agent coder} SM. In the class name generation task, we compare the \emph{inspection} SM, which introduces an additional \emph{inspect model} composite state, to the \emph{MIG} approach. In the question answering task, we examine the \emph{routing} SM and \emph{ReAct} SM.  We also evaluate the \emph{planning} SM that further enforces a predefined operation sequence commonly used in this task to avoid wasting LLM invocations.

\begin{figure*}
    \centering
    \includegraphics[width=0.32\textwidth]{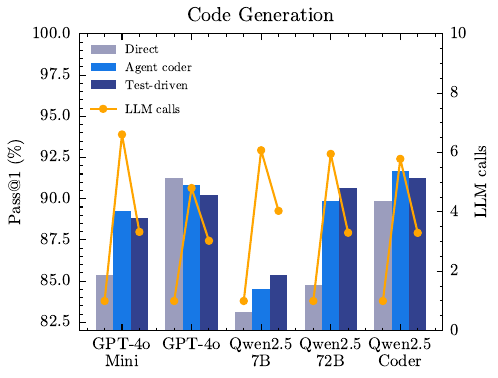}
    \includegraphics[width=0.32\textwidth]{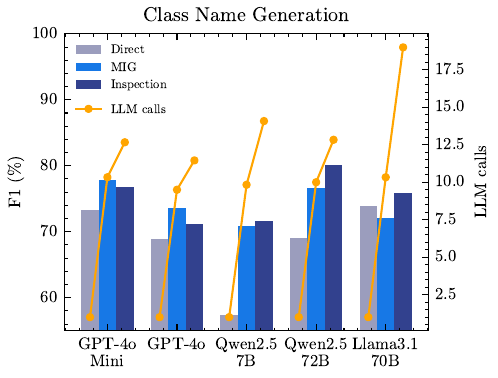}
    \includegraphics[width=0.32\textwidth]{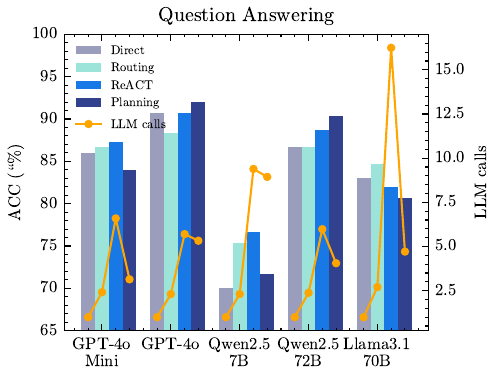}
    \caption{Comparison of different SM designs and the \direct{} method with respect to performance and number of LLM calls}
    \label{fig:rq2}
\end{figure*}

\paragraph{Impact on performance} 
\autoref{fig:rq2} presents the performance of various SM configurations across different use cases. Overall, all SM-based approaches generally outperform \direct{}, although the specific performance varies. For the code generation task, the \emph{test-driven} SM achieves comparable performance to the \emph{agent coder} SM, even surpassing it with Qwen2.5-7B and Qwen2.5 72B, indicating that repeatedly generating test cases may be unnecessary for the code generation scenario, particularly with Qwen models.

In the class name generation task, introducing the additional inspection state in the \emph{Inspection} SM enhances performance over the \emph{MIG} SM for both Qwen and Llama LLMs; however, it slightly degrades the performance for GPT-4o variants. Nevertheless, the combination of the \emph{Inspection} SM with Qwen2.5-72B yields the highest overall performance.

For the question answering task, we examine two distinct SM modeling strategies. While the \emph{routing} SM outperforms the \direct{} approach in three out of five cases, separating planning and information extraction in the \emph{ReAct} SM results in further performance gains in four out of five LLMs. Additionally, enforcing a predefined sequence of operations in the \emph{planning} SM enhances performance specifically for the two strongest LLMs tested in this task: GPT-4o and Qwen2.5 72B. We hypothesize that less capable LLMs might lack sufficient usage of predefined operation order (by consistently firing the \textit{continue} transitions), making the \emph{planning} SM less helpful.

\begin{tcolorbox}[boxsep=-1mm] 
   \textbf{RQ2.1.} 
   The SMs notably impact \approach{} performance, varying depending on the specific LLM size and LLM architecture. For tasks with well-defined best practices (e.g., code generation, class name generation), the impact seems to vary for different LLM families, while for tasks with less defined best practices (e.g., question answering), the impact seems to depend more on LLM size.
\end{tcolorbox}

\paragraph{Impact on cost} 
The orange lines in \autoref{fig:rq2} indicate the cost in terms of the average number of LLM calls required by each SM design. Across all tested LLMs and tasks, a clear trend emerges: SMs designed explicitly to minimize LLM calls indeed result in a reduced number of invocations. Specifically, the \emph{test-driven} SM achieves approximately 50\% fewer LLM calls on average compared to the \emph{agent coder} SM. Similarly, the pre-defined action orders in the \emph{planning} SM also reduce the number of LLM calls compared to the \emph{ReAct} SM, with a particularly significant reduction observed for Llama3.1 70B. Conversely, the additional \emph{inspect model} state in the \emph{Inspection} SM increases the number of LLM calls relative to the \emph{MIG} SM, since this state explicitly prompts the LLM to inspect outputs and potentially revert to previous states.

\begin{tcolorbox}[boxsep=-1mm] 
    \textbf{RQ2.2.}
    The configuration of SMs can influence the number of LLM calls in \approach{}. Generally, configurations explicitly designed to reduce LLM calls achieve notable reductions in cost. 
\end{tcolorbox}

\subsection{Discussion} 
\noindent \textbf{Applicability.} 
In RQ1, we demonstrate that integrating SMs into \approach{} consistently enhances LLM performance across three tasks, with the most pronounced improvements for code generation and class name generation. These tasks notably have clear human best practices, which effectively guide the LLM by systematically decomposing the tasks. In contrast, the question-answering task is more general and lacks well-defined best practices, resulting in relatively modest performance gains. We suspect that integrating SMs using \approach{} is more effective for tasks with established best practices, especially for use cases lacking enough training data (e.g., class name generation).


\noindent \textbf{Impact of state machine configuration.} 
In RQ2, we observe that the SM design significantly influences LLM performance, although this influence varies according to the LLM family and size. Due to this experimental nature, the separation of SM design from action implementation in \approach{} is especially valuable, enabling rapid experiments and optimization of different SM designs without altering underlying implementations.

Furthermore, SMs designed to optimize the number of LLM calls consistently achieve reductions across all tested tasks, indicating more predictable costs when using \approach{}. Importantly, these optimized SMs maintain performance comparable to unoptimized SMs, showing that cost savings do not come at the expense of effectiveness. This balance is particularly beneficial when using strong LLMs from an API, where computational costs can be substantial. We believe the explicit structure of SMs enables engineers to more accurately predict and manage LLM execution costs, potentially integrating them effectively with traditional model-checking techniques on SMs \cite{alur1998model}.

\subsection{Threats to validity}

\noindent\textbf{Internal validity.}
LLM outputs may be non-deterministic depending on the configuration and hardware. To mitigate this issue, we use a low sampling temperature (0.01) for all experiments and average results across three independent runs. Additionally, the performance of LLMs can vary based on prompting techniques. To minimize this variation, we consistently followed established techniques, such as chain-of-thought prompting~\cite{wei2022chain} and few-shot prompting~\cite{brown2020language-gpt3} for different approaches compared in the experiments.

\noindent\textbf{External validity.}
The impact of SMs could vary across different domains and tasks. In this paper, we evaluate SMs on three distinct use cases. However, the observed impacts may differ for other domains. We leave the exploration of SMs' impacts on other tasks and domains as future work. 

\noindent\textbf{Construct validity.}
The evaluation metrics adopted in this paper may raise a threat. For the \pass{} metric, passing all tests does not necessarily guarantee the correctness of the generated functions. Full correctness evaluation may require complex analysis of generated code. Nevertheless, this metric is widely recognized and used in existing research~\cite{abukhalaf2023codex}. For the class name generation and question-answering tasks, we likewise selected the most commonly used metrics~\cite{chen2023automated,johnson2017clevr}.
\section{Related Work}

\noindent
\textbf{MDE for LLMs.}
Applying MDE principles to enhance LLM applications is an emerging research area. Clarisó et al.~\cite{clariso2023model} introduce \texttt{Impromptu}, a domain-specific language (DSL) designed for defining prompts. Several frameworks \cite{cuadrado2024automating,zhang2025chainbuddy} also use DSL to define conversational agents with LLMs. For testing LLMs, Morales et al.~\cite{morales2024dsl} present \texttt{LangBite}, a model-driven approach for specifying ethical requirements and automating the testing of ethical biases in LLMs. 

\approach{} also uses a model-driven approach to optimize the use of LLMs. 
While methods like \texttt{Impromptu} primarily focus on the specific prompt, \approach{} works at a higher level of abstraction that constrains the flow of LLM-based applications.





\noindent
\textbf{LLMs for MDE.}
The use of LLMs for automating MDE tasks has emerged as a prominent research area~\cite{di2025use}. Notably, these models have been effectively used for model completion or recommendation~\cite{chaaben2023towards}, model query generation~\cite{abukhalaf2023codex,lopez2024text2vql}, domain model generation~\cite{camara2023assessment,chen2023automated}, among others~\cite{di2025use}.

Most approaches using LLMs for automating MDE tasks resemble the \textit{direct} approach, typically involving a single prompt or a linear sequence of prompts. For example, Chen et al.~\cite{chen2023automated} evaluate various LLMs for domain model generation using few-shot prompting.
Similarly, Abukhalaf et al.~\cite{abukhalaf2023codex} use different prompting strategies to generate OCL queries.  

The use of LLMs in these approaches are more similar to the \direct{} baselines used in this paper (for which SM integration shows benefit in the majority of the cases), we believe that, with well-designed SMs tailored for each scenario, \approach{} could be used to further enhance these MDE tasks.

\noindent
\textbf{Structured workflow for LLMs.}
Many approaches~\cite{shinn2024reflexion,yao2023react,lewis2020retrieval,langgraph} have been proposed to improve the performance of LLMs by using a structured workflow. 
The Reflexion framework~\cite{shinn2024reflexion} enhances LLM agents through self-feedback. 
In the ReAct framework~\cite{yao2023react}, LLMs generate reasoning traces and task-specific actions. 
The retrieval-augmented generation (RAG) using various tools has also been explored through various approaches~\cite{lewis2020retrieval, asai2023self, jiang2023active, khattab2022demonstrate}. 
To better manage external tools and enhance modularity, researchers have explored multi-agent systems~\cite{wu2024autogen, hong2024metagpt, liang2023encouraging, 10645429}. Specifically, Wu et al.~\cite{wu2024autogen} introduce AutoGen, which uses multiple customizable and interactive LLM agents to collaboratively accomplish tasks.


\approach{} can also be seen as a way to integrate structured workflows with LLMs. Unlike previous methods, where workflows are either fixed to specific use cases or tightly coupled with the implementation, \approach{} proposes a general approach to define workflows and can be used to implement existing approaches such as RAG and ReACT. 


\noindent
\textbf{State machines in LLMs.}
Relatively few research efforts have focused on integrating LLMs with state machine-like workflows~\cite{wu2024stateflow, liu2023smot}. Wu et al.~\cite{wu2024stateflow} propose StateFlow, a novel LLM-based task-solving framework that models complex task-solving processes as state machines, where transitions between states are governed by heuristic rules or LLM-driven decisions, with actions performed within each state.  Another notable effort is the State Machine of Thoughts~\cite{liu2023smot}, which uses a state machine to track experiences from previous reasoning trajectories. Langchain~\cite{langgraph} and similar frameworks~\cite{crewai2025,n8n2025,haystack2025,wu2024autogen} build agentic applications with LLMs using SM–like workflows or pipelines. However, these frameworks typically treat the workflow as pre-compiled and static, making it difficult to modify the workflow dynamically at runtime.


Compared to existing approaches, \approach{} introduces several improvements: (1) it uses a hierarchical SM to support more modular structure definition; (2) it decouples SM design from action implementation, enabling better modularity; (3) it supports various policies for navigating the SM, including both LLM and rule-based approaches; and (4) it treats SMs as data, allowing them to be dynamically updated.







\section{Conclusion}
\label{sec:conclusion}

In this paper, we propose \approach{}, a model-driven framework designed to enhance the performance of LLMs on complex tasks using domain-specific best practices with hierarchical SMs.
By structuring execution through SMs, \approach{} provides more flexible and fine-grained control over the behavior of LLM-based applications. 
We demonstrate the applicability and effectiveness of \approach{} across diverse tasks.
The results show that, 
although LLM performance is generally improved by SMs, it is highly influenced by the SM design.
While the SM designs impact task performance differently depending on the LLM, specially designed SMs reduce the execution cost while maintaining a similar performance. Such result makes the separation of SM definition and action implementation in \approach{} particularly useful to support rapid experiments. 

In future work, we plan to integrate different textual languages for SMs to simplify the design process. We also intend to explore applications of \approach{} in different domains beyond the three use cases. Additionally, investigating how \approach{} can be combined with complementary techniques, such as reinforcement learning-based policies, may further improve its performance and broaden its applicability.

\bibliographystyle{IEEEtran}
\bibliography{reference}

\end{document}